\pdfoutput=1

\documentclass[11pt]{article}

\usepackage[final]{acl}

\usepackage{times}
\usepackage{latexsym}

\usepackage[T1]{fontenc}

\usepackage[utf8]{inputenc}

\usepackage{microtype}

\usepackage{inconsolata}

\usepackage{graphicx}

\usepackage{tikz}
\usetikzlibrary{positioning}
\usepackage{amssymb}
\usepackage{amsmath}

\usepackage{url}
\usepackage{booktabs, multirow} 
\usepackage{soul}
\usepackage{graphics}
\usepackage[export]{adjustbox}
\usepackage{subfig}
\usepackage{enumitem}

\title{Aligning Pre-trained Models for Spoken Language Translation}

\author{Šimon Sedláček, Santosh Kesiraju, Alexander Polok, Jan Černocký \\
        Speech@FIT, Brno University of Technology, Czechia \\
        \texttt{\{isedlacek,kesiraju,ipoloka,cernocky\}@fit.vut.cz}}

\newcommand{\howto}[0]{\texttt{How2}}
\newcommand{\val}[0]{\texttt{val}}
\newcommand{\dev}[0]{\texttt{test}}
\newcommand{\dmodel}[0]{$d_\text{model}$}

\begin{document}

\maketitle
\begin{abstract}
This paper investigates a~novel approach to end-to-end speech translation (ST) based on aligning frozen pre-trained automatic speech recognition (ASR) and machine translation (MT) models via a~small connector module (Q-Former, our Subsampler-Transformer Encoder). This connector bridges the gap between the speech and text modalities, transforming ASR encoder embeddings into the latent representation space of the MT encoder while being the only part of the system optimized during training. Experiments are conducted on the How2 English-Portuguese dataset as we investigate the alignment approach in a~small-scale scenario focusing on ST. While keeping the size of the connector module constant and small in comparison (<\thinspace5\% of the size of the larger aligned models), increasing the size and capability of the foundation ASR and MT models universally improves translation results. We also find that the connectors can serve as domain adapters for the foundation MT models, significantly improving translation performance in the aligned ST setting. We conclude that this approach represents a~viable and scalable approach to training end-to-end ST systems.

\end{abstract}

\section{Introduction}
The task of speech translation (ST) is typically addressed by using either a~composite cascade ST system built from pre-trained automatic speech recognition (ASR) and machine translation (MT) models, or by training an end-to-end model, typically based on the transformer~\cite{transformer} architecture. However, both approaches have their caveats. Cascade systems can sometimes suffer from error accumulation due to potential domain and vocabulary mismatch of the foundation models, amplified by first transcribing the input utterance and only subsequently translating it. While end-to-end ST systems alleviate such issues, training them from scratch requires more ST data, often employing several pre-training stages and techniques involving transfer learning and additional model supervision, to obtain better results \cite{espnet_st, kesiraju_low}. 

This work attempts to strike a~middle-ground between the two ST approaches, aiming at leveraging frozen pre-trained ASR and MT models joined by a~small connector module, bridging the gap between their respective representation spaces.

\subsection{Related works}
A trend for this paradigm was set by \cite{blip2}, where a~connector network called Q-Former was used to map abstract image representations from a~frozen image encoder into the word embedding space of a~large language model (LLM), generating text annotations. During training, the foundation models remain frozen and only the connector is trained, decreasing the cost of the training process considerably. Other works have subsequently adopted a~similar approach to address different cross-modal tasks \cite{xllm, instruct_blip, videollama, macaw_llm, flamingo, panda_gpt, chen2023videollm}.

The same model alignment paradigm has also been adopted for ASR and other speech-oriented tasks. In \cite{qformer_asr}, the Q-Former is used to join large ASR encoders and LLMs for the ASR task. Similarly, a~linear-projection connector with either a~convolutional downsampling or CTC compression frontend was used in \cite{ctc_integration} for the same purpose. In \cite{wang2023slm}, a~large-scale multi-task training approach was used to enhance the capabilities of an instruction-tuned LLM to understand speech inputs, utilizing a~transformer connector with stochastic downsampling.

While these works all follow a~similar approach, training the connector network, the focus is on ASR or other general large-scale speech-language tasks. In this work, we adopt the alignment approach on a~much smaller scale with a~sole focus on ST. For ST, utilizing a~Q-Former-like connector network to bridge the gap between the speech and text modalities of frozen ASR and MT models is attractive for a~few reasons. As opposed to training a~conventional end-to-end ST model, the connector network can be much smaller with much fewer parameters. Moreover, even if the aligned pre-trained models are very powerful, the connector only has to learn an approximate mapping between their respective hidden representation spaces. Since only the connector parameters are optimized, training runs are shorter and require fewer computational resources, in contrast to conventional end-to-end models of similar sizes.

\subsection{Findings and contributions}
\setlist{nosep}
\begin{itemize}
    \item The presented alignment framework (Section~\ref{sec:alignment_archs}) is a~viable and generic approach to solving speech translation (Sections~\ref{sec:arch_eval},~\ref{sec:scaling}).
    \item We find that our STE connector (Section~\ref{sec:ste}) is superior to the Q-Former for our ST alignment scenario due its variable-length mapping capability, as determining an appropriate number of the Q-Former queries is difficult in comparison (Section~\ref{sec:qformer_vs_ste}).
    \item Scaling up the aligned ASR and MT models leads to universally better speech translation results, while the size of the connector network can remain constant, and relatively small (Section~\ref{sec:scaling}).
    \item The connector networks can serve as domain adapters, significantly improving translation performance for scenarios, where the aligned MT models are out-of-domain (Section~\ref{sec:domain_adapters}).
    \item We also study the effectiveness of the proposed framework under simulated low-resource scenarios (Section~\ref{sec:low_resource}).
\end{itemize}

\section{Alignment architectures}
\label{sec:alignment_archs}
We experiment with two general alignment architectures, differing in the configuration of the frozen MT model, as shown in Figure~\ref{fig:ecd_eced}. For both architectures, only the encoder part of the ASR model is used. Both architectures are trained using the standard cross-entropy loss objective at the output of the MT decoder.

\subsection{ECD}
The first alignment architecture (Figure~\ref{fig:ecd_eced}A) -- \emph{Encoder-Connector-Decoder} (ECD) -- consists of a~frozen ASR encoder, which extracts hidden audio representations from the source speech audio. These representations are then passed to a~connector module, whose outputs are further projected to the dimension of the frozen MT decoder via a~fully-connected layer -- the connector is used to align the ASR encoder output embedding space with the output embedding space of the MT encoder, essentially overtaking the responsibilities of the removed MT encoder, while simultaneously adapting to the speech modality\footnote{We find that if the connector and MT encoder architectures match, initializing the connector weights with the original MT encoder marginally speeds up convergence. No performance gains were observed.}.

\subsection{ECED}
\begin{figure*}[t]
    \centering
    \includegraphics[scale=0.9]{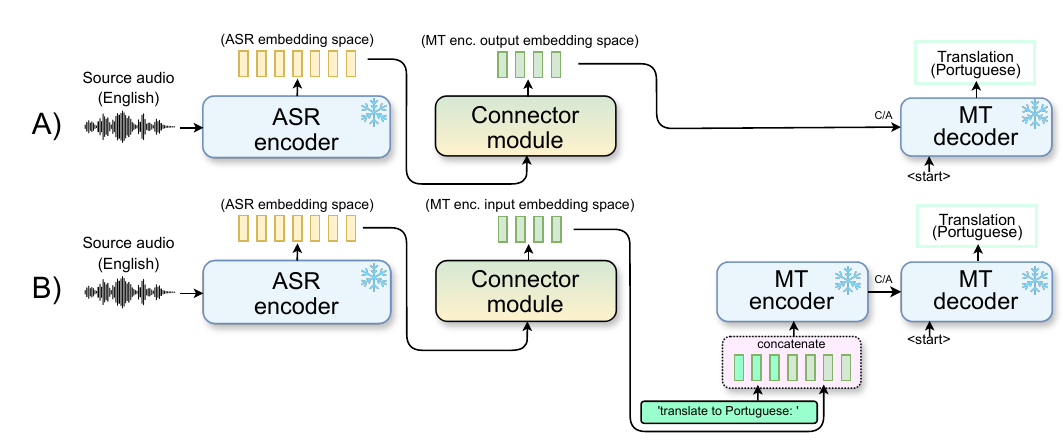}
    \caption{Diagram of the ECD (A) and the ECED (B) alignment architectures for ST. Modules annotated with the `*` symbol are frozen. In the ECD scenario, the connector outputs are passed directly to the cross-attention (C/A) connection of the MT decoder. For the ECED architecture, the connector outputs are directly injected past the input embedding layer of the MT encoder. Additional task prompt word embeddings can be prepended to the connector outputs before entering the MT encoder.}
    
    \label{fig:ecd_eced}
\end{figure*}

In the second \emph{Encoder-Connector-Encoder-Decoder} (ECED) setting (Figure~\ref{fig:ecd_eced}B), the connector module projects the speech embeddings into the input word embedding space of the MT encoder. Here, the modality gap between the output ASR embeddings and the input MT encoder text embeddings should be narrower to the ECD case, as the connector only has to propagate the raw textual information contained in the embeddings to the MT encoder. a~similar approach was explored in~\cite{wang2023slm}, where aligning the speech encoder with a~language model from the T5~\cite{t5} family allowed the authors to leverage the original multi-task and reasoning capabilities of the foundation multi-lingual language model, only enhanced in that the final model could also process speech inputs. This encompasses first taking the instruction prompt (`translate to Portuguese:'), embedding it with the MT encoder embedding layer, and prepending these instruction embeddings to the connector module outputs.

\section{Connector modules}
\label{sec:connectors}

This section presents the two connector modules used in our alignment experiments: the Q-Former and our STE connector.

\subsection{Q-Former}
\begin{figure}[h]
    \centering
    \hspace{-3em}
    \includegraphics[scale=0.9]{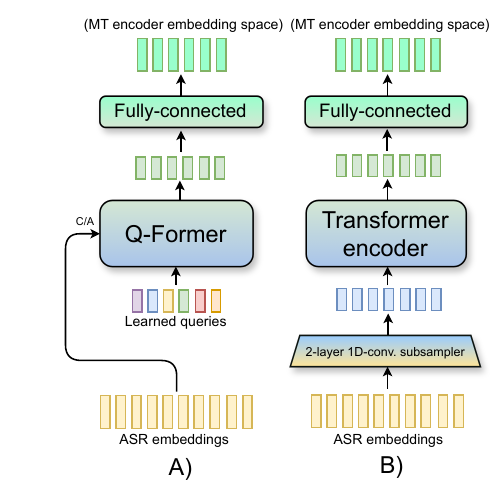}
    \caption{Diagram of the two connector modules. Variant A) is the Q-Former, variant B) is the STE connector.}
    \label{fig:connectors}
\end{figure}

The Q-Former \cite{blip2} (Figure~\ref{fig:connectors}A) is a~simple transformer model, which uses a~sequence of trainable \emph{queries} $\mathbf{Q}\in\mathbb{R}^{d_q \times n_q}$ as input ($d_q$ is the query dimension, $n_q$ is the number of queries used, which is a~hyperparameter). These queries interact with the supplied speech embedding sequence $\mathbf{S}\in\mathbb{R}^{d_s \times n_s}$ via cross-attention, extracting relevant information and transforming it into the MT model representation space. The Q-Former therefore performs a~variable to fixed-length sequence mapping, where the information density in the output query sequence $\mathbf{Q}'$ is directly correlated to the number of queries used $n_q$. Apart from the fixed-length mapping performed by the Q-Former, it should also be noted that the self-attention layer in the first decoder block operates only on the raw queries, providing no value in the transformation process. The Q-Former has previously been adopted in the ASR context \cite{qformer_asr}, therefore we treat it as a~baseline connector in our experiments.

\subsection{STE connector}
\label{sec:ste}
Additionally, we experiment with another connector variant inspired by~\cite{wang2023slm, ctc_integration} (Figure~\ref{fig:connectors}B). This variant addresses the fixed-length mapping of the Q-Former by introducing a~convolutional subsampling layer in replacement of the queries and cross-attention connection, and we, therefore, refer to it as the `STE` connector (Subsampler-Transformer Encoder).

The subsampling layer is used to reduce the time dimension granularity of the supplied ASR embeddings to better match the granularity and information density of the token sequences processed by the MT model. In~\cite{ctc_integration}, CTC compression and convolutional downsampling are explored, evaluating the convolutional approach as superior. In~\cite{wang2023slm}, similar subsampling is done by stochastically discarding a~fourth of the ASR embeddings. Our subsampling module is a~2-layer stack of 1D convolutions, reducing the length of the input speech embedding sequence $\mathbf{S}\in\mathbb{R}^{d_s \times n_s}$ by a~factor of 4, and simultaneously projecting the embeddings into the hidden size $d_c$ of the connector:\vspace{-3mm}
\begin{equation}
\mathbf{S}'\in\mathbb{R}^{d_c\times n_s/4} = \text{Conv}(\mathbf{S}). \vspace{-3mm}
\end{equation}
The subsampler was inspired by the 1D conv. frontend used for ASR systems in~\cite{fairseq_s2t}. The subsampler is then followed by a~stack of transformer encoder blocks, and finally, the output embeddings are projected into the MT model dimension $d_t$ by a~fully-connected layer.

\section{Experimental setup}
We conduct our experiments within the Hugging Face Transformers \cite{wolf-etal-2020-transformers} ecosystem, which allows us to easily access and reuse various pre-trained off-the-shelf ASR and MT models for our experiments. All training runs were done on a~single Nvidia RTX A5000 24GB GPU, with most alignment runs lasting around 10 hours on average, depending on the sizes of the foundation models.

\subsection{Data and evaluation}
All models in this work are trained and evaluated using the \howto{} dataset \cite{how2}. \howto{} is a~multi-modal dataset of English instructional videos, containing a~smaller 300-hour audio subset with crowd-sourced Portuguese translations, for which the two validation and test partitions (\val{}, \dev{}) both total just under 4 hours of speech data. \howto{} is a~commonly used standard benchmark dataset for MT and ST systems~\cite{iwslt2019-how2, espnet_st, joint_transformers_slt, kesiraju_low}. As the data is relatively clean, it is considered to be appropriate for studying and evaluating various approaches for ST.

For our ASR models, we report normalized WER. Our translation models are evaluated using 4-gram BLEU\footnote{\tiny{\texttt{case:mixed|tok:13a|ngram:4|smooth:False}}}~\cite{bleu} and chrF2\footnote{\tiny{\texttt{case:mixed|char\_o:6|word\_o:0|$\beta$:2|whtspc:False|eps\_smooth:False}}} scores \cite{popovic-2015-chrf} as implemented in the Hugging Face Evaluate v0.4.0 library.

\section{Foundation and baseline models}
\label{sec:foundation_models}
To better ground our results, we take inspiration from ESPnet \cite{espnet_st} \howto{}  recipes\footnote{\url{https://github.com/espnet/espnet/tree/master/egs/how2}} and use the best-performing transformer models as reference points when establishing our ASR, MT and ST baseline systems. We use these reference models to evaluate our alignment methods in a~more restricted in-domain scenario on the \howto{} dataset, and subsequently also choose other, more powerful off-the-shelf pre-trained ASR and MT models to evaluate the alignment approach in a~more general scenario, exploring the dimensionality and domain adaptation capabilities of the connectors.

\subsection{ASR models}
\label{sec:asr_models}
For our reference ASR system (referred to as E-Branchformer small), we train a~small 12-layer E-Branchformer \cite{ebranchformer} model with a~6-layer transformer decoder for English ASR on the \howto{} dataset. The model uses 4 attention heads, $d_\text{model}$ of 256, and a~lower-cased unigram~\cite{unigram} sub-word vocabulary of size 5000 with punctuation removed. In accordance to the ESPnet reference, an additional CTC objective~\cite{joint_attention_ctc_transformer} with a~weight of 0.3 is used alongside the standard cross-entropy loss. We use 80-dimensional log-mel-filterbanks as features. Speed perturbation in factors [0.9, 1.0, 1.1] as well as SpecAugment \cite{specaugment} in the `LD' strategy were applied during training. The final model achieves 12.2\% WER on the \dev{} set, as shown in Table~\ref{tab:baseline_asr_systems}.

\begin{table}[]
\centering
\resizebox{\columnwidth}{!}{
\begin{tabular}{lcrc}
\toprule
\multirow{2}{*}{Model} & WER & \multirow{2}{*}{Params.} &  \multirow{2}{*}{$d_\text{model}$} \\ 
    & \dev{} & & \\ \midrule
CTC/attn. E-Branch. small   &  12.2    & 38.5\tiny{M}  & 256 \\
CTC/attn. E-Branch. medium &  11.7    & 174\tiny{M}    & 512 \\
Whisper-small.en &              7.9    & 242\tiny{M}   & 768  \\ \midrule
ESPnet CTC/attn. transformer &  13.0   & 30\tiny{M}    & 256     \\ \bottomrule
\end{tabular}
}
\caption{WER performance of the foundation ASR systems evaluated on the \howto{} \dev{} set. The E-Branchformer small and ESPnet reference models were trained solely using \howto{} data, the E-Branchformer medium was not. The ESPnet system is included for reference.}
\label{tab:baseline_asr_systems}
\end{table}

For the off-the-shelf models, we choose two models that have not previously seen any of the \howto{} data. First, we use a~scaled-up version of our baseline CTC/attn. ASR system -- 16-layer E-Branchformer model with an 8-layer decoder and \dmodel{} of 512. This model was trained on 6k hours of English data, described in Appendix~\ref{sec:appendix_ebr_medium} along with evaluation results. As shown in Table~\ref{tab:baseline_asr_systems}, the model achieves 11.7\% WER on the \dev{} set without fine-tuning, but we hope to leverage its representation-building capabilities in our alignment experiments. It will be further referred to as E-Branchformer medium. Lastly, we use OpenAI Whisper-small.en \cite{whisper} as our last and most powerful foundation ASR model, achieving 7.9\% WER on the \dev{} set.

\subsection{MT models}
\label{sec:mt_models}
Our baseline MT system adopts the MarianMT~\cite{marianmt} transformer implementation available in Hugging Face Transformers. Again, following the ESPnet \howto{} recipes, both the encoder and decoder consist of 6 layers with 4 attention heads, $d_\text{model}$ of 256, and intermediate layer size of 2048. The source and target word embedding layers were untied in accordance with~\cite{espnet_st}. The system was trained in a~true-cased to true-cased manner with both the source and target BPE-based~\cite{BPE} vocabularies containing 8000 word units. As shown in Table~\ref{tab:baseline_mt_systems}, the model achieves 57 BLEU on the \howto{} \dev{} set, matching the results of ESPnet.

\begin{table}[h]
\centering
\resizebox{0.9\columnwidth}{!}{
\begin{tabular}{lccrc}
\toprule
\multirow{2}{*}{Model} & \multicolumn{2}{c}{BLEU} & \multirow{2}{*}{Params.} &  \multirow{2}{*}{$d_\text{model}$} \\ 
                & \val{} & \dev{} & & \\ \midrule
MarianMT small         & 57.9 & 57.0              & 21.6\tiny{M}    & 256 \\
T5 En-Pt               & 40.0 & 38.8              & 223\tiny{M}     & 768 \\ \bottomrule
\end{tabular}
}
\caption{BLEU performance of the foundation MT systems, evaluated on the \howto{} \dev{} set. The MarianMT model was trained on \howto{}, the T5 model is out-of-domain on the \howto{} dataset.}
\label{tab:baseline_mt_systems}
\end{table}

Additionally, we select an En-Pt MT model based on the T5~\cite{t5} architecture from~\cite{t5_port}. The model was based on the English T5-base checkpoint, which was first pre-trained on Portuguese language~\cite{ptt5}, and then fine-tuned for English to Portuguese translation on a~5M English-Portuguese sentence subset of ParaCrawl~\cite{espla-etal-2019-paracrawl}, as well as domain-specific data in preparation for the WMT19 and WMT20 biomedical translation tasks. The encoder and decoder consist of 12 layers with a~\dmodel{} of 768. This MT model is out-of-domain on \howto{}, achieving `only' 38.8 BLEU on the \dev{} set. However, we chose it in hopes of being able to leverage its Portuguese text-generation capabilities in our alignment experiments. The model is freely available on Hugging Face\footnote{\url{https://huggingface.co/unicamp-dl/translation-en-pt-t5}} and will be further referred to as simply the T5 model.

\subsection{ST baselines}
\label{sec:st_baselines}

From our E-Branchformer small and MarianMT models trained on \howto{}, we construct two baseline ST systems. First, we once again adopt the ESPnet approach on \howto{}, where we use the ASR encoder and MT decoder of the foundation systems to initialize the encoder and decoder of the ST system. The vocabulary of the decoder is the same true-cased BPE vocabulary used by the MarianMT model. The system is then trained for a~maximum of 40 epochs with a~learning rate of $1\text{e}^{-3}$, 10000 warm-up steps, batch size of 128, and early stopping to avoid overfitting. SpecAugment is no longer used, however, we keep the speed perturbation with the same factors of [0.9, 1.0, 1.1]. The encoder is kept frozen for the first 8 epochs of training. This baseline system achieves 45.6 and 45.2 BLEU on the \val{} and \dev{} \howto{} subsets, respectively, which is on par with the results of the reference ESPnet system, as shown in Table~\ref{tab:slt_baselines}.

This E2E system represents the more robust, although more costly solution to the ST task, obtaining good performance using a~small model, which employs only one decoding step and does not suffer from any domain mismatch. However, optimizing all the model parameters is necessary to get the best results, leading to bigger resource consumption if the model were to scale up.

\begin{table}[h]
\centering
\resizebox{\columnwidth}{!}{
\begin{tabular}{lcc}
\toprule
\multirow{2}{*}{Baseline ST system} & \multicolumn{2}{c}{BLEU}   \\
                   & \val{} & \dev{} \\ \midrule
E-Branchformer E2E (ASR + MT init.)                     & 45.6                      & 45.2           \\
Cascade (Ebr. ASR $\rightarrow$ \texttt{tc} $\rightarrow$ MarianMT)                  & 40.9                      & 40.4           \\
\midrule
ESPnet transformer reference                     & -                        & 45.7         \\
\bottomrule
\end{tabular}
}
\caption{Comparison of both trained and reference ST systems. The ESPnet reference is an E2E transformer system, utilizing the same ASR+MT initialization approach as our E-Branchformer end-to-end system.}
\label{tab:slt_baselines}
\end{table}

Table~\ref{tab:slt_baselines} also shows a~cascade ST system we construct from the same E-Branchformer small and MarianMT models. Because the vocabularies of the models are mismatched both in size and in casing, we take a~pre-trained T5-based English casing and punctuation restoration model from Hugging Face\footnote{\url{https://huggingface.co/SJ-Ray/Re-Punctuate}} and use it to process the ASR transcriptions before translating them with the MT model. Our cascade system represents a~\emph{naive} but cheap solution to the ST task, as no parameters have to be tuned. However, the system has to perform three decoding steps and the overall performance suffers from error accumulation. The model only achieves 40.4 BLEU on the \dev{} set, though the chrF2 scores reported later in Table~\ref{tab:arch_comparison} paint the model in a~more positive light.

\section{Alignment architecture evaluation}
\label{sec:arch_eval}
In the first set of alignment experiments, we evaluate the two alignment architecture variants (ECD, ECED) in conjunction with both of the connector network types (Q-Former, STE). These experiments aim to primarily determine the viability of the aligning approach to training ST systems as a~whole, juxtaposing them against the baseline methods. On top of that, we test the effects of using different numbers of layers in the connector. Unless specified otherwise, all aligned models are trained for a~maximum of 70 epochs with 15000 warm-up steps, a~batch size of 128, and a~peak learning rate of $2e^{-4}$. We adopt early stopping to avoid overfitting and apply speed perturbation with the factors of [0.9, 1.0, 1.1]. Other than the layer count, the connector network parameters are kept constant, with a~\dmodel{} of 256, 4 attention heads, and intermediate layer size of 2048, keeping the Q-Former query count at 100. The results are shown in Table~\ref{tab:arch_comparison}, Appendix~\ref{sec:appendix_table} then contains the full Table~\ref{tab:arch_comparison_appendix}.

\begin{table}[h]
\centering
\resizebox{\columnwidth}{!}{%
\scriptsize
\begin{tabular}{lccrccccc}\toprule
\multirow{2}{*}{Arch.} &\multirow{2}{*}{C} &\multirow{2}{*}{\shortstack{L}} &\multirow{2}{*}{\shortstack{\#P}} &\multicolumn{2}{c}{\howto{} BLEU} & \multicolumn{2}{c}{\howto{} chrF2}\\\cmidrule{5-6} \cmidrule{7-8}
& & & &\val & \dev & \val &\dev \\\midrule
ECD &Q   &6 &9.6\tiny{M} &44.0 &43.9 & 65.0 & 64.8 \\
 &STE &6 &10.7\tiny{M} &\textbf{45.0} &\textbf{44.8} & \textbf{66.0} & \textbf{65.8} \\
 &STE &4 &7.9\tiny{M} &44.1 &44.4 & 65.3 & 65.3 \\ \midrule
ECED &Q   &6 &9.6\tiny{M} &44.1 &44.2 & 65.3 & 65.2 \\
     &STE &6 &10.7\tiny{M} &\textbf{44.7} &\textbf{44.8} & \textbf{65.8} & \textbf{65.9} \\ \midrule
\multicolumn{3}{l}{E-Branch. E2E} &38.5\tiny{M} & \textbf{45.6} & \textbf{45.2} & \textbf{66.6} & \textbf{66.1} \\ 
\multicolumn{3}{l}{Cascade} & - & 40.9 & 40.4 & 64.4 & 64.1 \\ 
\bottomrule
\end{tabular}%
}
\caption{Alignment architecture comparison for the E-Branchformer small and MarianMT foundation models. Both connectors have \dmodel{} of 256, the Q-Former uses 100 queries. Headers C, L, and \#P denote the connector type, connector layers and number of trainable parameters, respectively.}
\label{tab:arch_comparison}
\end{table}

\subsection{ECD evaluation}
\label{sec:ecd_eval}
As is shown in the first half of Table~\ref{tab:arch_comparison} the STE connector outperforms the Q-Former by a~significant margin, almost matching the performance of the baseline E2E system, achieving 45.0 and 44.8 BLEU on the \val{} and \dev{} sets, respectively. It could be argued, that the STE connector perhaps benefits too much from the parameters added by the convolutional subsampler, however, even the 4-layer STE configuration still outperforms the 6-layer Q-Former, suggesting that the problem lies probably in the way the Q-Former represents and extracts information from the input speech embeddings. This is further analyzed in Section~\ref{sec:qformer_vs_ste}. We also find that the STE training process is more stable and `well-behaved` than the one of the Q-Former, with faster convergence and fewer spikes in validation metrics during training.

\subsection{ECED evaluation}
\label{sec:eced_eval}
For the ECED architecture, the performance falloff is not as stark when decreasing the number of connector layers, compared to ECD. The best model once again uses the 6-layer STE connector, with results on par with the equivalent ECD configuration, achieving 44.7 and 44.8 BLEU on the \val{} and \dev{} sets, respectively. The Q-Former maintains better results in contrast to the ECD architecture, reinforcing the hypothesis that the mapping problem for the ECED architecture is a~simpler one to that of ECD.

Concluding for both architectures, most of the aligned models outperform our cascade baseline. Though none of them surpass the E2E baseline system, comparable performance is obtained only tuning a~small module with less than a~quarter of the parameters of the E2E system. This indicates that there is potential in the alignment approach to ST, and we further explore what can be achieved by scaling up the foundation ASR and MT models in Section~\ref{sec:scaling}.

\subsection{Evaluating the connectors on different input lengths}
\label{sec:qformer_vs_ste}

As the STE connector outperforms the Q-Former in previous experiments, we argue that this stems from the variable to fixed-length mapping performed by the Q-Former, directly defined by the number of queries used. Contrary to \cite{qformer_asr}, we find that both choosing this number to be too low (40\thinspace-\thinspace60) or too high (128+) comes with performance degradation, as the Q-Former either does not have enough queries to represent the necessary information or is unable to distribute it among the queries effectively. We find that for every foundation model configuration, there is a~sweet spot for the number of queries used, which has to be experimentally determined. On the contrary, the STE connector output length is always monotonously correlated to the amount of represented information, reducing the risk of propagating noise to the downstream MT model.

\begin{figure}
    \centering
    \includegraphics[width=0.9\columnwidth]{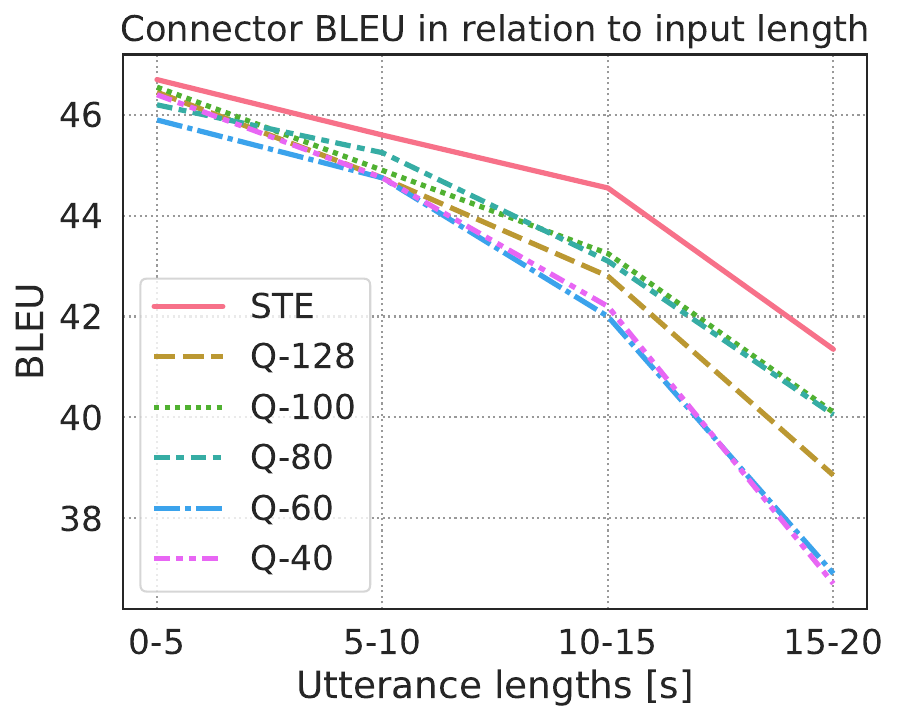}
    \caption{Performance study of the STE connector and the Q-Former (with varying numbers of queries) in relation to input utterance lengths. The final BLEU score is obtained by averaging BLEUs computed for both the \val{} and \dev{} \howto{} sets.}
    \label{fig:q_vs_ste_lengths}
\end{figure}

To better demonstrate the Q-Former performance degradation, we train five models with 40, 60, 80, 100, and 128 queries in the E-Branchformer small/MarianMT ECD configuration and compare them to the best STE model from Section~\ref{sec:ecd_eval}. For this purpose, we partition the \val{} and \dev{} sets into four subsets by utterance lengths: 0\thinspace-\thinspace5, 5\thinspace-\thinspace10, 10\thinspace-\thinspace15, and 15\thinspace-\thinspace20 seconds. Because the majority of \howto{} concentrates in the 0 to 10-second range, the two long-utterance splits are considerably smaller than the two shorter ones. However, we think the results show a~clear trend beyond any potential errors stemming from the smaller amounts of test data (though the effect could probably be mitigated by adding longer utterances to the training set). To obtain the final BLEU value for each sub-split, the scores obtained for the \val{} and \dev{} splits are averaged to better illustrate the overall performance decline trend of each evaluated system. The comparison is shown in Figure~\ref{fig:q_vs_ste_lengths}.

\section{Foundation model scaling}
\label{sec:scaling}
We investigate the prospect of aligning different off-the-shelf (possibly out-of-domain) pre-trained ASR and MT models, aiming to improve ST performance by increasing the capability of the aligned models. Ideally, the alignment process would overcome any domain mismatch that would appear if the same systems were used as parts of a~cascade ST system\footnote{We do not build a~cascade system using the T5 model, as its performance on \howto{} is impaired (Table~\ref{tab:baseline_mt_systems}) by domain mismatch, leaving no hope for good cascade ST results.}.

The size of the connector network is kept the same as in previous experiments (6 layers, \dmodel{} of 256), allowing for shorter and more stable training runs. It could be argued that a~connector with a~\dmodel{} closer to that of the foundation models would yield better performance, however, we argue that such behavior is expected and the more interesting finding is that the connector can, in fact, be much smaller than either of the foundation models, while still bringing ST performance improvements when scaling up the ASR and MT models.

\begin{table*}[!htp]\centering
\resizebox{1.9\columnwidth}{!}{%
\scriptsize
\begin{tabular}{lcccccrcccc}\toprule
\multirow{2}{*}{ASR enc.} &\multirow{2}{*}{Connector} &\multirow{2}{*}{\shortstack{Connector \\ layers}} &\multirow{2}{*}{Queries} &\multirow{2}{*}{MT enc.} &\multirow{2}{*}{MT dec.} &\multirow{2}{*}{\shortstack{Trainable \\ parameters}} &\multicolumn{2}{c}{\howto{} BLEU} & \multicolumn{2}{c}{\howto{} chrF2} \\\cmidrule{8-9} \cmidrule{10-11}
& & & & & & &\val &\dev &\val &\dev \\\midrule
Ebr. medium &Q &6 &128 &- &Marian 6l &10.4\tiny{M} &45.6 &45.7 & 66.5 & 66.5 \\
Ebr. medium &Q &6 &128 &- &T5 12l &10.5\tiny{M} &46.8 &47.5 & 67.8 & 68.2 \\
Ebr. small &Q &6 &128 &- &T5 12l &9.7\tiny{M} &44.5 &44.4 & 65.9 & 65.4\vspace{1mm} \\
Ebr. medium &STE &6 &- &- &Marian 6l &12.0\tiny{M} &46.0 &46.3 & 66.7 & 66.7 \\
Ebr. medium &STE &6 &- &- &T5 12l &12.0\tiny{M} &47.8 &48.5 & 68.8 & 68.9 \\
Ebr. small &STE &6 &- &- &T5 12l &10.7\tiny{M} &45.4 &45.6 & 67.0 & 66.7 \vspace{1mm}\\
Ebr. medium &STE &6 &- &T5 12l &T5 12l &12.0\tiny{M} &47.5 &48.0 & 68.6 & 69.0 \\
Ebr. small &STE &6 &- &T5 12l &T5 12l &10.7\tiny{M} &45.2 &45.7 & 66.9 & 67.2 \vspace{1mm}\\
Whisper small &Q &6 &100 &- &T5 12l &11.3\tiny{M} &47.4 &47.6 & 66.4 & 66.3 \\
Whisper small &STE &6 &- &- &T5 12l &13.3\tiny{M} &\textbf{48.2} & \textbf{48.9} & \textbf{68.9} & \textbf{69.2} \\\midrule
\multicolumn{6}{l}{E-Branchformer E2E baseline} &38.5\tiny{M} & 45.6 & 45.2 & 66.6 & 66.1 \\ 
\multicolumn{6}{l}{Cascade baseline} & - & 40.9 & 40.4 & 64.4 & 64.1 \\ 
\bottomrule
\end{tabular}%
}
\caption{Aligned ST system performance with different combinations of scaled-up foundation ASR and MT models.}
\label{tab:alignment_scaling}
\end{table*}

All systems are trained with the same training hyper-parameters as in previous experiments except for the two Whisper-small runs, where we change the batch size to 48 with 3 gradient accumulation steps because of memory constraints. We train most models in this section in the ECD alignment configuration, with both the Q-Former and STE connectors. The results are shown in Table~\ref{tab:alignment_scaling}.

Once again, the STE connector outperforms the Q-Former across all possible foundation model configurations. The results also suggest that increasing the size and capability of the ASR encoder might generally yield a~bigger performance improvement than when using a~bigger MT decoder. However, more experiments should be done to confirm this hypothesis, as the chosen T5 model is perhaps not the fairest reference point to draw such conclusions. Of course, utilizing both bigger ASR encoders and bigger MT decoders yields the best translation results, as is best demonstrated by the experiments we performed with the Whisper-small.en model. In combination with the same STE connector\footnote{Both the Whisper encoder and the T5 decoder have a~\dmodel{} of 768, the connector still has a~\dmodel{} of 256.}, this system yields the best performance among all trained models, achieving 48.9 BLEU on the \dev{} set.

\subsection{Connector modules as domain adapters}
\label{sec:domain_adapters}

Interestingly, the BLEU scores achieved by the T5 system in the ECD aligned scenario with the Whisper model far supersede the raw MT scores on \howto{} from Section~\ref{sec:mt_models}, bringing an improvement of over 9 BLEU points on the \dev{} set compared to the base T5 MT scenario\footnote{The original T5 MT score on the \dev{} set was only 38.8 BLEU.}. We argue that this is due to the connector network being able to serve as a~domain adapter, reinforcing the notion that the foundation ASR and MT models might not require fine-tuning on the available ST data before alignment.

We also train two ECED systems with the T5 model, as one might argue that a~big part of the domain adaptation process is actually cutting off the T5 encoder. To match the original operational context of the T5 model, the tokenized version of the task prompt `translate English to Portuguese:' is first embedded via the T5 encoder input embedding layer and then prepended to the connector output embedding sequence, similarly to~\cite{wang2023slm}, and as shown in Figure~\ref{fig:ecd_eced}.

The expectation would be that if the connector only tried to match the text embedding representation of the translated sentence at the input of the MT encoder, the domain adaptation would fail, as there would be no distinction between the aligned scenario and the basic MT scenario in the same domain. However (lines 7\thinspace-\thinspace8 in Table~\ref{tab:alignment_scaling}), both T5 models with either ASR encoder almost match the performances of their ECD counterparts. This suggests that the connector is doing something more abstract, \emph{steering} the behavior of the MT model in a~more nuanced way than just passing it transformed ASR embeddings with textual information, allowing the MT encoder to properly translate the original sentence, without the out-of-domain effect becoming an issue.

\section{Low resource scenario}
\label{sec:low_resource}
To better understand the ST data requirements for successful alignment, we train two ECED systems with the same 6-layer STE connector on three low-resource simulation \howto{} splits, as similar scenarios are not discussed in the related works.

\begin{table}[!htp]\centering
\resizebox{\columnwidth}{!}{%
\scriptsize
\begin{tabular}{lcccc}\toprule
Model &\multicolumn{1}{c}{17h} &\multicolumn{1}{c}{51h} &\multicolumn{1}{c}{153h} &\multicolumn{1}{c}{300h} \\\midrule
E-Branch. small + MarianMT &25.0 &38.3 &43.3 &44.8 \\
E-Branch. medium + T5      &33.5 &40.6 &45.4 &48.0 \\
\bottomrule
\end{tabular}%
}
\caption{ECED/STE system BLEU performance on the \dev{} set when trained on different volumes of data.}
\label{tab:low_resource}
\end{table}

As shown in Table~\ref{tab:low_resource}, the training is reasonably successful even with the 153-hour \howto{} split. Results for the 17 and 51-hour splits show potential for future work, as we still observe a~clear performance gain from utilizing more powerful foundation models. Namely, for the 51-hour split, the ST result of the T5 model still exceeds its performance in the base MT task on the \dev{} set.

\section{Conclusions}

In this paper, we explored an approach of aligning pre-trained frozen ASR and MT models for ST. Our experiments show that the alignment approach is a~viable generic framework for establishing new end-to-end speech ST. We find that scaling up the foundation ASR and MT models leads to universally better ST results, while the size of the connector network can remain constant, and relatively small. Our STE connector outperforms the Q-Former due to performance degradation on longer input sequences and the difficulty of choosing an appropriate number of queries for the given task. The connectors also demonstrate good domain adaptation capabilities, improving translation performance by more than 9 BLEU points on the \dev{} \howto{} set in case of the out-of-domain T5 model with respect to its base MT performance on the \dev{} set.

\section{Limitations}

We identify some limitations of the presented alignment ST approach. First, although initial experiments (Section~\ref{sec:low_resource}) are promising, further investigation is required to study the effectiveness of the proposed framework in low-resource scenarios. We think that reducing the amount of ST data needed for successfully performing the alignment procedure should be a~primary concern for future work so that this alignment paradigm can become viable even for truly low-resource scenarios.

Second, perhaps unsurprisingly, the small size of the connector network ultimately starts to negatively impact ST performance if the foundation model \dmodel{} exceeds a~certain threshold. We attempt to use a~233M parameter En-Pt MT model\footnote{\url{https://huggingface.co/Helsinki-NLP/opus-mt-tc-big-en-pt}} from the OPUS project \cite{tiedemann-thottingal-2020-opus, tiedemann-2020-tatoeba} in the aligned scenario. This model achieves 58.2 BLEU on the \dev{} set for the MT task without fine-tuning, suggesting good prospects for the aligned ST scenario. However, we find that using the same 6-layer STE connector with \dmodel{} of 256, the training takes longer than for other models to converge, ultimately yielding `only' 47.3 BLEU on the \dev{} set in conjunction with Whisper in the ECD setting. As this MT model has a~\dmodel{} of 1024, we suspect that the slightly subpar performance is likely caused by the hidden size mismatch between the connector and the MT model. More experiments need to be conducted to confirm this and we leave those for future work.

\bibliography{acl_latex}

\begin{thebibliography}{40}
\providecommand{\natexlab}[1]{#1}

\bibitem[{Alayrac et~al.(2022)Alayrac, Donahue, Luc, Miech, Barr, Hasson, Lenc, Mensch, Millican, Reynolds, Ring, Rutherford, Cabi, Han, Gong, Samangooei, Monteiro, Menick, Borgeaud, Brock, Nematzadeh, Sharifzadeh, Binkowski, Barreira, Vinyals, Zisserman, and Simonyan}]{flamingo}
Jean{-}Baptiste Alayrac, Jeff Donahue, Pauline Luc, Antoine Miech, Iain Barr, Yana Hasson, Karel Lenc, Arthur Mensch, Katherine Millican, Malcolm Reynolds, Roman Ring, Eliza Rutherford, Serkan Cabi, Tengda Han, Zhitao Gong, Sina Samangooei, Marianne Monteiro, Jacob~L. Menick, Sebastian Borgeaud, Andy Brock, Aida Nematzadeh, Sahand Sharifzadeh, Mikolaj Binkowski, Ricardo Barreira, Oriol Vinyals, Andrew Zisserman, and Kar{\'{e}}n Simonyan. 2022.
\newblock \href {http://papers.nips.cc/paper\_files/paper/2022/hash/960a172bc7fbf0177ccccbb411a7d800-Abstract-Conference.html} {Flamingo: a visual language model for few-shot learning}.
\newblock In \emph{Advances in Neural Information Processing Systems 35: Annual Conference on Neural Information Processing Systems 2022, NeurIPS 2022, New Orleans, LA, USA, November 28 - December 9, 2022}.

\bibitem[{Ardila et~al.(2020)Ardila, Branson, Davis, Kohler, Meyer, Henretty, Morais, Saunders, Tyers, and Weber}]{ardila_common_2020}
Rosana Ardila, Megan Branson, Kelly Davis, Michael Kohler, Josh Meyer, Michael Henretty, Reuben Morais, Lindsay Saunders, Francis Tyers, and Gregor Weber. 2020.
\newblock \href {https://aclanthology.org/2020.lrec-1.520} {Common {Voice}: {A} {Massively}-{Multilingual} {Speech} {Corpus}}.
\newblock In \emph{Proceedings of the {Twelfth} {Language} {Resources} and {Evaluation} {Conference}}, pages 4218--4222, Marseille, France. European Language Resources Association.

\bibitem[{Carmo et~al.(2020)Carmo, Piau, Campiotti, Nogueira, and Lotufo}]{ptt5}
Diedre Carmo, Marcos Piau, Israel Campiotti, Rodrigo Nogueira, and Roberto Lotufo. 2020.
\newblock Ptt5: Pretraining and validating the t5 model on brazilian portuguese data.
\newblock \emph{arXiv preprint arXiv:2008.09144}.

\bibitem[{Chen et~al.(2023{\natexlab{a}})Chen, Han, Zhao, Zhang, Shi, Xu, and Xu}]{xllm}
Feilong Chen, Minglun Han, Haozhi Zhao, Qingyang Zhang, Jing Shi, Shuang Xu, and Bo~Xu. 2023{\natexlab{a}}.
\newblock \href {https://arxiv.org/abs/2305.04160} {X-llm: Bootstrapping advanced large language models by treating multi-modalities as foreign languages}.
\newblock \emph{Preprint}, arXiv:2305.04160.

\bibitem[{Chen et~al.(2023{\natexlab{b}})Chen, Zheng, Wang, Xu, Huang, Pan, Wang, Wang, Qiao, Lu, and Wang}]{chen2023videollm}
Guo Chen, Yin-Dong Zheng, Jiahao Wang, Jilan Xu, Yifei Huang, Junting Pan, Yi~Wang, Yali Wang, Yu~Qiao, Tong Lu, and Limin Wang. 2023{\natexlab{b}}.
\newblock \href {https://arxiv.org/abs/2305.13292} {Videollm: Modeling video sequence with large language models}.
\newblock \emph{Preprint}, arXiv:2305.13292.

\bibitem[{Dai et~al.(2023)Dai, Li, Li, Tiong, Zhao, Wang, Li, Fung, and Hoi}]{instruct_blip}
Wenliang Dai, Junnan Li, Dongxu Li, Anthony Meng~Huat Tiong, Junqi Zhao, Weisheng Wang, Boyang Li, Pascale Fung, and Steven C.~H. Hoi. 2023.
\newblock \href {http://papers.nips.cc/paper\_files/paper/2023/hash/9a6a435e75419a836fe47ab6793623e6-Abstract-Conference.html} {Instructblip: Towards general-purpose vision-language models with instruction tuning}.
\newblock In \emph{Advances in Neural Information Processing Systems 36: Annual Conference on Neural Information Processing Systems 2023, NeurIPS 2023, New Orleans, LA, USA, December 10 - 16, 2023}.

\bibitem[{Espl{\`a} et~al.(2019)Espl{\`a}, Forcada, Ram{\'\i}rez-S{\'a}nchez, and Hoang}]{espla-etal-2019-paracrawl}
Miquel Espl{\`a}, Mikel Forcada, Gema Ram{\'\i}rez-S{\'a}nchez, and Hieu Hoang. 2019.
\newblock \href {https://aclanthology.org/W19-6721} {{P}ara{C}rawl: Web-scale parallel corpora for the languages of the {EU}}.
\newblock In \emph{Proceedings of Machine Translation Summit XVII: Translator, Project and User Tracks}, pages 118--119, Dublin, Ireland. European Association for Machine Translation.

\bibitem[{Gage(1994)}]{BPE}
Philip Gage. 1994.
\newblock A new algorithm for data compression.
\newblock \emph{C Users J.}, 12(2):23–38.

\bibitem[{Godfrey et~al.(1992)Godfrey, Holliman, and McDaniel}]{godfrey_switchboard_1992}
J.J. Godfrey, E.C. Holliman, and J.~McDaniel. 1992.
\newblock \href {https://doi.org/10.1109/ICASSP.1992.225858} {{SWITCHBOARD}: telephone speech corpus for research and development}.
\newblock In \emph{[{Proceedings}] {ICASSP}-92: 1992 {IEEE} {International} {Conference} on {Acoustics}, {Speech}, and {Signal} {Processing}}, volume~1, pages 517--520 vol.1.

\bibitem[{Hernandez et~al.(2018)Hernandez, Nguyen, Ghannay, Tomashenko, and Estève}]{hernandez_ted-lium_2018}
Francois Hernandez, Vincent Nguyen, Sahar Ghannay, Natalia Tomashenko, and Yannick Estève. 2018.
\newblock \href {https://doi.org/10.1007/978-3-319-99579-3_21} {{TED}-{LIUM} 3: {Twice} as {Much} {Data} and {Corpus} {Repartition} for {Experiments} on {Speaker} {Adaptation}: 20th {International} {Conference}, {SPECOM} 2018, {Leipzig}, {Germany}, {September} 18–22, 2018, {Proceedings}}.
\newblock pages 198--208.

\bibitem[{Hono et~al.(2023)Hono, Mitsuda, Zhao, Mitsui, Wakatsuki, and Sawada}]{ctc_integration}
Yukiya Hono, Koh Mitsuda, Tianyu Zhao, Kentaro Mitsui, Toshiaki Wakatsuki, and Kei Sawada. 2023.
\newblock \href {https://arxiv.org/abs/2312.03668} {An integration of pre-trained speech and language models for end-to-end speech recognition}.
\newblock \emph{Preprint}, arXiv:2312.03668.

\bibitem[{Inaguma et~al.(2020)Inaguma, Kiyono, Duh, Karita, Yalta, Hayashi, and Watanabe}]{espnet_st}
Hirofumi Inaguma, Shun Kiyono, Kevin Duh, Shigeki Karita, Nelson Yalta, Tomoki Hayashi, and Shinji Watanabe. 2020.
\newblock \href {https://doi.org/10.18653/v1/2020.acl-demos.34} {{ESP}net-{ST}: All-in-one speech translation toolkit}.
\newblock In \emph{Proceedings of the 58th Annual Meeting of the Association for Computational Linguistics: System Demonstrations}, pages 302--311, Online. Association for Computational Linguistics.

\bibitem[{Junczys-Dowmunt et~al.(2018)Junczys-Dowmunt, Grundkiewicz, Dwojak, Hoang, Heafield, Neckermann, Seide, Germann, Aji, Bogoychev, Martins, and Birch}]{marianmt}
Marcin Junczys-Dowmunt, Roman Grundkiewicz, Tomasz Dwojak, Hieu Hoang, Kenneth Heafield, Tom Neckermann, Frank Seide, Ulrich Germann, Alham~Fikri Aji, Nikolay Bogoychev, Andr{\'e} F.~T. Martins, and Alexandra Birch. 2018.
\newblock \href {https://doi.org/10.18653/v1/P18-4020} {{M}arian: Fast neural machine translation in {C}++}.
\newblock In \emph{Proceedings of {ACL} 2018, System Demonstrations}, pages 116--121, Melbourne, Australia. Association for Computational Linguistics.

\bibitem[{Karita et~al.(2019)Karita, Soplin, Watanabe, Delcroix, Ogawa, and Nakatani}]{joint_attention_ctc_transformer}
Shigeki Karita, Nelson Enrique~Yalta Soplin, Shinji Watanabe, Marc Delcroix, Atsunori Ogawa, and Tomohiro Nakatani. 2019.
\newblock \href {https://doi.org/10.21437/Interspeech.2019-1938} {{Improving Transformer-Based End-to-End Speech Recognition with Connectionist Temporal Classification and Language Model Integration}}.
\newblock In \emph{Proc. Interspeech 2019}, pages 1408--1412.

\bibitem[{Kesiraju et~al.(2023)Kesiraju, Sarvaš, Pavlíček, Macaire, and Ciuba}]{kesiraju_low}
Santosh Kesiraju, Marek Sarvaš, Tomáš Pavlíček, Cécile Macaire, and Alejandro Ciuba. 2023.
\newblock \href {https://doi.org/10.21437/interspeech.2023-2506} {Strategies for improving low resource speech to text translation relying on pre-trained asr models}.
\newblock In \emph{INTERSPEECH 2023}. ISCA.

\bibitem[{Kim et~al.(2022)Kim, Wu, Peng, Pan, Sridhar, Han, and Watanabe}]{ebranchformer}
Kwangyoun Kim, Felix Wu, Yifan Peng, Jing Pan, Prashant Sridhar, Kyu~Jeong Han, and Shinji Watanabe. 2022.
\newblock \href {https://doi.org/10.1109/SLT54892.2023.10022656} {E-branchformer: Branchformer with enhanced merging for speech recognition}.
\newblock In \emph{{IEEE} Spoken Language Technology Workshop, {SLT} 2022, Doha, Qatar, January 9-12, 2023}, pages 84--91. {IEEE}.

\bibitem[{Kudo(2018)}]{unigram}
Taku Kudo. 2018.
\newblock \href {https://doi.org/10.18653/v1/P18-1007} {Subword regularization: Improving neural network translation models with multiple subword candidates}.
\newblock In \emph{Proceedings of the 56th Annual Meeting of the Association for Computational Linguistics (Volume 1: Long Papers)}, pages 66--75, Melbourne, Australia. Association for Computational Linguistics.

\bibitem[{Li et~al.(2023)Li, Li, Savarese, and Hoi}]{blip2}
Junnan Li, Dongxu Li, Silvio Savarese, and Steven C.~H. Hoi. 2023.
\newblock \href {https://proceedings.mlr.press/v202/li23q.html} {{BLIP-2:} bootstrapping language-image pre-training with frozen image encoders and large language models}.
\newblock In \emph{International Conference on Machine Learning, {ICML} 2023, 23-29 July 2023, Honolulu, Hawaii, {USA}}, volume 202 of \emph{Proceedings of Machine Learning Research}, pages 19730--19742. {PMLR}.

\bibitem[{Lopes et~al.(2020)Lopes, Nogueira, Lotufo, and Pedrini}]{t5_port}
Alexandre Lopes, Rodrigo Nogueira, Roberto Lotufo, and Helio Pedrini. 2020.
\newblock \href {https://www.aclweb.org/anthology/2020.wmt-1.90} {Lite training strategies for {P}ortuguese-{E}nglish and {E}nglish-{P}ortuguese translation}.
\newblock In \emph{Proceedings of the Fifth Conference on Machine Translation}, pages 833--840, Online. Association for Computational Linguistics.

\bibitem[{Lyu et~al.(2023)Lyu, Wu, Wang, Huang, Liu, Du, Shi, and Tu}]{macaw_llm}
Chenyang Lyu, Minghao Wu, Longyue Wang, Xinting Huang, Bingshuai Liu, Zefeng Du, Shuming Shi, and Zhaopeng Tu. 2023.
\newblock \href {https://arxiv.org/abs/2306.09093} {Macaw-llm: Multi-modal language modeling with image, audio, video, and text integration}.
\newblock \emph{Preprint}, arXiv:2306.09093.

\bibitem[{Niehues et~al.(2019)Niehues, Cattoni, St{\"u}ker, Negri, Turchi, Ha, Salesky, Sanabria, Barrault, Specia, and Federico}]{iwslt2019-how2}
Jan Niehues, Rolando Cattoni, Sebastian St{\"u}ker, Matteo Negri, Marco Turchi, Thanh-Le Ha, Elizabeth Salesky, Ramon Sanabria, Loic Barrault, Lucia Specia, and Marcello Federico. 2019.
\newblock \href {https://aclanthology.org/2019.iwslt-1.1} {The {IWSLT} 2019 evaluation campaign}.
\newblock In \emph{Proceedings of the 16th International Conference on Spoken Language Translation}, Hong Kong. Association for Computational Linguistics.

\bibitem[{Panayotov et~al.(2015)Panayotov, Chen, Povey, and Khudanpur}]{panayotov_librispeech_2015}
Vassil Panayotov, Guoguo Chen, Daniel Povey, and Sanjeev Khudanpur. 2015.
\newblock \href {https://doi.org/10.1109/ICASSP.2015.7178964} {Librispeech: {An} {ASR} corpus based on public domain audio books}.
\newblock In \emph{2015 {IEEE} {International} {Conference} on {Acoustics}, {Speech} and {Signal} {Processing} ({ICASSP})}, pages 5206--5210.
\newblock ISSN: 2379-190X.

\bibitem[{Papineni et~al.(2002)Papineni, Roukos, Ward, and Zhu}]{bleu}
Kishore Papineni, Salim Roukos, Todd Ward, and Wei-Jing Zhu. 2002.
\newblock \href {https://doi.org/10.3115/1073083.1073135} {{B}leu: a method for automatic evaluation of machine translation}.
\newblock In \emph{Proceedings of the 40th Annual Meeting of the Association for Computational Linguistics}, pages 311--318, Philadelphia, Pennsylvania, USA. Association for Computational Linguistics.

\bibitem[{Park et~al.(2019)Park, Chan, Zhang, Chiu, Zoph, Cubuk, and Le}]{specaugment}
Daniel~S. Park, William Chan, Yu~Zhang, Chung-Cheng Chiu, Barret Zoph, Ekin~D. Cubuk, and Quoc~V. Le. 2019.
\newblock \href {https://doi.org/10.21437/interspeech.2019-2680} {Specaugment: A simple data augmentation method for automatic speech recognition}.
\newblock In \emph{Interspeech 2019}. ISCA.

\bibitem[{Paul and Baker(1992)}]{paul_design_1992}
Douglas~B. Paul and Janet~M. Baker. 1992.
\newblock \href {https://aclanthology.org/H92-1073} {The {Design} for the {Wall} {Street} {Journal}-based {CSR} {Corpus}}.
\newblock In \emph{Speech and {Natural} {Language}: {Proceedings} of a {Workshop} {Held} at {Harriman}, {New} {York}, {February} 23-26, 1992}.

\bibitem[{Popovi{\'c}(2015)}]{popovic-2015-chrf}
Maja Popovi{\'c}. 2015.
\newblock \href {https://doi.org/10.18653/v1/W15-3049} {chr{F}: character n-gram {F}-score for automatic {MT} evaluation}.
\newblock In \emph{Proceedings of the Tenth Workshop on Statistical Machine Translation}, pages 392--395, Lisbon, Portugal. Association for Computational Linguistics.

\bibitem[{Radford et~al.(2023)Radford, Kim, Xu, Brockman, McLeavey, and Sutskever}]{whisper}
Alec Radford, Jong~Wook Kim, Tao Xu, Greg Brockman, Christine McLeavey, and Ilya Sutskever. 2023.
\newblock \href {https://proceedings.mlr.press/v202/radford23a.html} {Robust speech recognition via large-scale weak supervision}.
\newblock In \emph{International Conference on Machine Learning, {ICML} 2023, 23-29 July 2023, Honolulu, Hawaii, {USA}}, volume 202 of \emph{Proceedings of Machine Learning Research}, pages 28492--28518. {PMLR}.

\bibitem[{Raffel et~al.(2020)Raffel, Shazeer, Roberts, Lee, Narang, Matena, Zhou, Li, and Liu}]{t5}
Colin Raffel, Noam Shazeer, Adam Roberts, Katherine Lee, Sharan Narang, Michael Matena, Yanqi Zhou, Wei Li, and Peter~J. Liu. 2020.
\newblock \href {http://jmlr.org/papers/v21/20-074.html} {Exploring the limits of transfer learning with a unified text-to-text transformer}.
\newblock \emph{J. Mach. Learn. Res.}, 21:140:1--140:67.

\bibitem[{Sanabria et~al.(2018)Sanabria, Caglayan, Palaskar, Elliott, Barrault, Specia, and Metze}]{how2}
Ramon Sanabria, Ozan Caglayan, Shruti Palaskar, Desmond Elliott, Lo\"ic Barrault, Lucia Specia, and Florian Metze. 2018.
\newblock \href {http://arxiv.org/abs/1811.00347} {{How2:} a large-scale dataset for multimodal language understanding}.
\newblock In \emph{Proceedings of the Workshop on Visually Grounded Interaction and Language (ViGIL)}. NeurIPS.

\bibitem[{Su et~al.(2023)Su, Lan, Li, Xu, Wang, and Cai}]{panda_gpt}
Yixuan Su, Tian Lan, Huayang Li, Jialu Xu, Yan Wang, and Deng Cai. 2023.
\newblock \href {https://aclanthology.org/2023.tllm-1.2} {{P}anda{GPT}: One model to instruction-follow them all}.
\newblock In \emph{Proceedings of the 1st Workshop on Taming Large Language Models: Controllability in the era of Interactive Assistants!}, pages 11--23, Prague, Czech Republic. Association for Computational Linguistics.

\bibitem[{Tiedemann(2020)}]{tiedemann-2020-tatoeba}
J{\"o}rg Tiedemann. 2020.
\newblock \href {https://aclanthology.org/2020.wmt-1.139} {The tatoeba translation challenge {--} realistic data sets for low resource and multilingual {MT}}.
\newblock In \emph{Proceedings of the Fifth Conference on Machine Translation}, pages 1174--1182, Online. Association for Computational Linguistics.

\bibitem[{Tiedemann and Thottingal(2020)}]{tiedemann-thottingal-2020-opus}
J{\"o}rg Tiedemann and Santhosh Thottingal. 2020.
\newblock \href {https://aclanthology.org/2020.eamt-1.61} {{OPUS}-{MT} {--} building open translation services for the world}.
\newblock In \emph{Proceedings of the 22nd Annual Conference of the European Association for Machine Translation}, pages 479--480, Lisboa, Portugal. European Association for Machine Translation.

\bibitem[{Vaswani et~al.(2017)Vaswani, Shazeer, Parmar, Uszkoreit, Jones, Gomez, Kaiser, and Polosukhin}]{transformer}
Ashish Vaswani, Noam Shazeer, Niki Parmar, Jakob Uszkoreit, Llion Jones, Aidan~N Gomez, \L~ukasz Kaiser, and Illia Polosukhin. 2017.
\newblock \href {https://proceedings.neurips.cc/paper_files/paper/2017/file/3f5ee243547dee91fbd053c1c4a845aa-Paper.pdf} {Attention is all you need}.
\newblock In \emph{Advances in Neural Information Processing Systems}, volume~30. Curran Associates, Inc.

\bibitem[{Vydana et~al.(2021)Vydana, Karafi\'{a}t, \v{Z}mol\'{i}kov\'{a}, Burget, and \v{C}ernock\'{y}}]{joint_transformers_slt}
K.~Hari Vydana, Martin Karafi\'{a}t, Kate\v{r}ina \v{Z}mol\'{i}kov\'{a}, Luk\'{a}\v{s} Burget, and Jan \v{C}ernock\'{y}. 2021.
\newblock \href {https://doi.org/10.1109/ICASSP39728.2021.9414159} {Jointly trained transformers models for spoken language translation}.
\newblock In \emph{ICASSP 2021 - 2021 IEEE International Conference on Acoustics, Speech and Signal Processing (ICASSP)}, pages 7513--7517. IEEE Signal Processing Society.

\bibitem[{Wang et~al.(2021)Wang, Riviere, Lee, Wu, Talnikar, Haziza, Williamson, Pino, and Dupoux}]{wang_voxpopuli_2021}
Changhan Wang, Morgane Riviere, Ann Lee, Anne Wu, Chaitanya Talnikar, Daniel Haziza, Mary Williamson, Juan Pino, and Emmanuel Dupoux. 2021.
\newblock \href {https://doi.org/10.18653/v1/2021.acl-long.80} {{VoxPopuli}: {A} {Large}-{Scale} {Multilingual} {Speech} {Corpus} for {Representation} {Learning}, {Semi}-{Supervised} {Learning} and {Interpretation}}.
\newblock In \emph{Proceedings of the 59th {Annual} {Meeting} of the {Association} for {Computational} {Linguistics} and the 11th {International} {Joint} {Conference} on {Natural} {Language} {Processing} ({Volume} 1: {Long} {Papers})}, pages 993--1003, Online. Association for Computational Linguistics.

\bibitem[{Wang et~al.(2020)Wang, Tang, Ma, Wu, Okhonko, and Pino}]{fairseq_s2t}
Changhan Wang, Yun Tang, Xutai Ma, Anne Wu, Dmytro Okhonko, and Juan Pino. 2020.
\newblock \href {https://aclanthology.org/2020.aacl-demo.6} {Fairseq {S}2{T}: Fast speech-to-text modeling with fairseq}.
\newblock In \emph{Proceedings of the 1st Conference of the Asia-Pacific Chapter of the Association for Computational Linguistics and the 10th International Joint Conference on Natural Language Processing: System Demonstrations}, pages 33--39, Suzhou, China. Association for Computational Linguistics.

\bibitem[{Wang et~al.(2023)Wang, Han, Shafran, Wu, Chiu, Cao, Chen, Zhang, Soltau, Rubenstein, Zilka, Yu, Pundak, Siddhartha, Schalkwyk, and Wu}]{wang2023slm}
Mingqiu Wang, Wei Han, Izhak Shafran, Zelin Wu, Chung{-}Cheng Chiu, Yuan Cao, Nanxin Chen, Yu~Zhang, Hagen Soltau, Paul~K. Rubenstein, Lukas Zilka, Dian Yu, Golan Pundak, Nikhil Siddhartha, Johan Schalkwyk, and Yonghui Wu. 2023.
\newblock \href {https://doi.org/10.1109/ASRU57964.2023.10389703} {{SLM:} bridge the thin gap between speech and text foundation models}.
\newblock In \emph{{IEEE} Automatic Speech Recognition and Understanding Workshop, {ASRU} 2023, Taipei, Taiwan, December 16-20, 2023}, pages 1--8. {IEEE}.

\bibitem[{Wolf et~al.(2020)Wolf, Debut, Sanh, Chaumond, Delangue, Moi, Cistac, Rault, Louf, Funtowicz, Davison, Shleifer, von Platen, Ma, Jernite, Plu, Xu, Le~Scao, Gugger, Drame, Lhoest, and Rush}]{wolf-etal-2020-transformers}
Thomas Wolf, Lysandre Debut, Victor Sanh, Julien Chaumond, Clement Delangue, Anthony Moi, Pierric Cistac, Tim Rault, Remi Louf, Morgan Funtowicz, Joe Davison, Sam Shleifer, Patrick von Platen, Clara Ma, Yacine Jernite, Julien Plu, Canwen Xu, Teven Le~Scao, Sylvain Gugger, Mariama Drame, Quentin Lhoest, and Alexander Rush. 2020.
\newblock \href {https://doi.org/10.18653/v1/2020.emnlp-demos.6} {Transformers: State-of-the-art natural language processing}.
\newblock In \emph{Proceedings of the 2020 Conference on Empirical Methods in Natural Language Processing: System Demonstrations}, pages 38--45, Online. Association for Computational Linguistics.

\bibitem[{Yu et~al.(2023)Yu, Tang, Sun, Chen, Tan, Li, Lu, Ma, and Zhang}]{qformer_asr}
Wenyi Yu, Changli Tang, Guangzhi Sun, Xianzhao Chen, Tian Tan, Wei Li, Lu~Lu, Zejun Ma, and Chao Zhang. 2023.
\newblock \href {https://arxiv.org/abs/2309.13963} {Connecting speech encoder and large language model for asr}.
\newblock \emph{Preprint}, arXiv:2309.13963.

\bibitem[{Zhang et~al.(2023)Zhang, Li, and Bing}]{videollama}
Hang Zhang, Xin Li, and Lidong Bing. 2023.
\newblock \href {https://doi.org/10.18653/v1/2023.emnlp-demo.49} {Video-{LL}a{MA}: An instruction-tuned audio-visual language model for video understanding}.
\newblock In \emph{Proceedings of the 2023 Conference on Empirical Methods in Natural Language Processing: System Demonstrations}, pages 543--553, Singapore. Association for Computational Linguistics.

\end{thebibliography}

\appendix

\section{E-Branchformer medium}
\label{sec:appendix_ebr_medium}

The E-Branchformer medium model was trained on a~6000-hour multi-domain English dataset comprised of the following datasets: Fisher (SWITCHBOARD)~\cite{godfrey_switchboard_1992},  WSJ~\cite{paul_design_1992}, Common Voice en 13~\cite{ardila_common_2020}, LibriSpeech~\cite{panayotov_librispeech_2015}, VoxPopuli~\cite{wang_voxpopuli_2021}, TED-LIUM~3~\cite{hernandez_ted-lium_2018}. Table~\ref{tab:ebr_results} shows the model performance when evaluated on these datasets as well as on \howto{}.

\begin{table}[h]
    \centering
    \resizebox{0.65\columnwidth}{!}{
    \begin{tabular}{lr} \toprule
    Dataset & WER \\ \midrule
    LibriSpeech (\texttt{test-clean}) & 2.5 \\
    LibriSpeech (\texttt{test-other}) & 5.6 \\
    TED-LIUM 3 &  6.3 \\
    VoxPopuli &  7.3 \\
    Common Voice en 13 &  12.1 \\
    How2 (\dev{}) &  11.7 \\
         \bottomrule
    \end{tabular}
    }
    \caption{E-Branchformer medium evaluation WERs on different datasets.}
    \label{tab:ebr_results}
\end{table}

\pagebreak
\section{Architecture evaluation full results}
\label{sec:appendix_table}

\begin{table}[h]
\centering
\resizebox{\columnwidth}{!}{%
\scriptsize
\begin{tabular}{lccrccccc}\toprule
\multirow{2}{*}{Arch.} &\multirow{2}{*}{C} &\multirow{2}{*}{\shortstack{L}} &\multirow{2}{*}{\shortstack{\#P}} &\multicolumn{2}{c}{\howto{} BLEU} & \multicolumn{2}{c}{\howto{} chrF2}\\\cmidrule{5-6} \cmidrule{7-8}
& & & &\val & \dev & \val &\dev \\\midrule
ECD &Q   &6 &9.6\tiny{M} &44.0 &43.9 & 65.0 & 64.8 \\
 &Q   &4 &6.4\tiny{M} &42.5 &42.6 & 63.9 & 63.7 \\
 &Q   &2 &3.2\tiny{M} &40.6 &40.2 & 61.9 & 62.3 \\
 &STE &6 &10.7\tiny{M} &\textbf{45.0} &\textbf{44.8} & \textbf{66.0} & \textbf{65.8} \\
 &STE &4 &7.9\tiny{M} &44.1 &44.4 & 65.3 & 65.3 \\
 &STE &2 &5.3\tiny{M} &42.8 &42.9 & 64.1 & 64.0 \\ \midrule
ECED &Q   &6 &9.6\tiny{M} &44.1 &44.2 & 65.3 & 65.2 \\
     &Q   &4 &6.4\tiny{M} &43.8 &43.7 & 65.1 & 64.8 \\
     &Q   &2 &3.2\tiny{M} &42.3 &41.5 & 63.6 & 62.9 \\
     &STE &6 &10.7\tiny{M} &\textbf{44.7} &\textbf{44.8} & \textbf{65.8} & \textbf{65.9} \\
     &STE &4 &8.1\tiny{M} &44.0 &44.1 & 65.2 & 65.2 \\
     &STE &2 &5.5\tiny{M} &43.0 &43.1 & 64.3 & 64.1 \\ \midrule 
\multicolumn{3}{l}{E-Branch. E2E} &38.5\tiny{M} & \textbf{45.6} & \textbf{45.2} & \textbf{66.6} & \textbf{66.1} \\ 
\multicolumn{3}{l}{Cascade} & - & 40.9 & 40.4 & 64.4 & 64.1 \\ 
\bottomrule
\end{tabular}%
}
\caption{Full alignment architecture comparison for the baseline E-Branchformer small and MarianMT foundation models. Both connectors have \dmodel{} of 256, the Q-Former uses 100 queries. Headers C, L, and \#P denote the connector type, connector layers, and number of trainable parameters, respectively.}
\label{tab:arch_comparison_appendix}
\end{table}

\end{document}